\documentclass{article}
\usepackage[main, final]{neurips_2026}
\makeatletter
\def\@noticestring{}
\def\@notice{}
\makeatother
\usepackage[utf8]{inputenc} 
\usepackage[T1]{fontenc}    
\usepackage{hyperref}      
\usepackage{url}            
\usepackage{booktabs}       
\usepackage{amsfonts}       
\usepackage{nicefrac}       
\usepackage{microtype}      
\usepackage{xcolor}   
\usepackage{amsmath}
\usepackage{graphicx}
\usepackage{booktabs}
\usepackage{multirow}
\usepackage[table]{xcolor}
\usepackage{tabularx}      

\title{Data Organization Matters in Multimodal Instruction Tuning: A Controlled Study of Capability Trade-offs}

\author{
Guowei Tang \\
School of Statistics \\
East China Normal University \\
\texttt{wwgTang2021@163.com}
}

\begin{document}

\maketitle

\begin{abstract}
Recent multimodal large language models (MLLMs) have achieved strong performance on general visual understanding, diagram and chart reasoning, and document-centric perception. However, these capabilities are typically acquired from heterogeneous supervision sources with substantially different task structures and learning demands, and the effect of how such data are temporally organized during training remains underexplored. In this work, we study whether data organization affects the capability trade-off among general understanding, structured reasoning, and fine-grained OCR/document understanding in multimodal instruction tuning. To isolate this factor, we adopt a controlled three-stage training framework in which the backbone, trainable modules, and optimization pipeline are fixed across all runs, and only the temporal arrangement of post-alignment supervision is varied. We compare four representative strategies: direct mixture, curriculum training, balanced sampling, and reverse curriculum. Experiments are conducted on a validation suite spanning general visual instruction following, diagram reasoning, chart reasoning, scene-text question answering, and document question answering, together with aggregated metrics for reasoning, OCR, and overall performance. We further analyze convergence speed, training stability. The results show that data organization is not a secondary implementation detail but a first-order design variable in multimodal adaptation. In particular, curriculum training yields the best overall trade-off and the strongest structured reasoning performance; balanced sampling is more favorable for OCR-oriented capability but sacrifices part of the broader capability balance; and reverse curriculum performs worst in both final performance and optimization stability. Training-dynamics analysis further suggests that establishing general understanding and reasoning before introducing OCR-intensive supervision leads to smoother optimization and faster convergence. These findings highlight data scheduling as an explicit design dimension for multimodal model adaptation and provide practical motivation for future work on adaptive curricula and finer-grained task organization.
\end{abstract}

\section{Introduction}
Recent multimodal large language models (MLLMs) have achieved strong performance across a wide range of vision--language tasks, including open-ended visual instruction following \cite{liu2023visual,dai2023instructblip}, diagram and chart reasoning \cite{hu2023paperowl,xia2024chartx}, and text-centric document understanding \cite{hu2024mplug}. Despite this progress, these capabilities are typically learned from heterogeneous supervision sources that differ substantially in task structure, perceptual granularity, and reasoning demand \cite{dai2023instructblip}. In practice, MLLMs are often trained on a combination of general visual instruction data, structured reasoning datasets, and OCR-intensive corpora \cite{dai2023instructblip,hu2024mplug}. However, although prior work has extensively studied model architectures and optimization strategies \cite{li2023blip2,dai2023instructblip}, the effect of \emph{how heterogeneous data are organized during training} remains underexplored.

This issue is important because different data sources emphasize different capabilities. General visual instruction data mainly supports broad semantic grounding and instruction-following ability \cite{liu2023visual,dai2023instructblip}. Diagram and chart datasets emphasize structured reasoning over visual layouts and symbolic relations \cite{kembhavi2016diagram,masry2022chartqa}. In contrast, OCR- and document-centric datasets require fine-grained perception of text tokens, local regions, and layout details \cite{singh2019textvqa,mathew2021docvqa}. While these objectives can be complementary, they may also compete for limited adaptation capacity during parameter-efficient and sequential multimodal tuning \cite{hu2022lora,ge2025dmole,chen2025sefe}. Consequently, the ordering and composition of training data may affect not only the final average performance, but also the balance among general understanding, structured reasoning, and fine-grained OCR ability \cite{ge2025dmole,chen2025sefe}.

In this work, we investigate whether the \emph{organization strategy of training data} influences the capability trade-off among \emph{general visual understanding, structured reasoning, and fine-grained OCR/document understanding} \cite{ge2025dmole,chen2025sefe}. To isolate this factor, we keep the training modules fixed across all experimental conditions. Specifically, the projector is trained alone during the initial alignment stage, following staged vision--language alignment practice \cite{li2023blip2,liu2023visual}; in subsequent stages, the projector and LLM LoRA are jointly optimized \cite{hu2022lora}; and the vision tower remains frozen throughout the entire training process, consistent with frozen-backbone multimodal adaptation pipelines \cite{li2023blip2}. This controlled design ensures that any observed performance difference can be attributed primarily to data organization rather than to changes in trainable modules.

We consider four representative data organization strategies. Our primary setting is a \emph{curriculum-style schedule}, which first emphasizes general visual instruction and structured reasoning, and then introduces OCR- and document-intensive supervision in a later stage \cite{bengio2009curriculum,ge2025dmole}. We compare this strategy with three alternatives: \emph{direct mixture}, which removes curriculum ordering and uses a fixed mixed distribution throughout the post-alignment stages; \emph{balanced sampling}, which enforces equal sampling probability across task sources; and \emph{reverse curriculum}, which prioritizes OCR-heavy data before shifting toward general understanding and structured reasoning. Importantly, these conditions are designed to maintain approximately matched total data exposure, so that the primary difference lies in the temporal organization of supervision rather than in the amount of training signal. This design allows us to directly examine whether training order influences the final capability trade-off, a question closely related to recent studies on continual and sequential multimodal instruction tuning \cite{ge2025dmole,chen2025sefe}.

To evaluate the impact of these strategies, we assess model performance on a validation suite spanning the target capability spectrum, including a general visual instruction validation set, diagram reasoning, chart reasoning, text-based visual question answering, and document visual question answering \cite{liu2023visual,kembhavi2016diagram,masry2022chartqa,singh2019textvqa,mathew2021docvqa}. We further report aggregated metrics for reasoning ability, OCR capability, and overall performance, and analyze convergence speed and training stability. Through this study, we aim to answer a practical question in multimodal instruction tuning: \emph{when the objective is to strike a balance among general understanding, reasoning ability, and OCR-intensive perception, does the training order matter? If so, under our setup, which data organization strategy delivers a better trade-off across these capabilities?} \cite{bengio2009curriculum,ge2025dmole,chen2025sefe}

Overall, the main contributions of this work are summarized as follows:
\begin{itemize}
    \item We formulate and systematically study the problem of how training data organization affects the balance among general visual understanding, structured reasoning, and fine-grained OCR/document understanding in multimodal instruction tuning.
    \item We design a controlled experimental framework that fixes the trainable modules and training protocol, enabling a fair comparison among curriculum training, direct mixture, balanced sampling, and reverse curriculum.
    \item We provide a comprehensive empirical analysis across task performance, convergence behavior, training stability, offering practical insights into data scheduling for multimodal model adaptation.
\end{itemize}

\section{Related Work}

\subsection{Multimodal Large Language Models and Visual Instruction Tuning}

Recent progress in multimodal large language models has largely been driven by the combination of strong pretrained language models and visual encoders, together with large-scale multimodal instruction tuning \cite{alayrac2022flamingo,li2023blip2,liu2023visual}. Early representative systems such as Flamingo established the effectiveness of bridging frozen vision and language backbones for few-shot multimodal learning \cite{alayrac2022flamingo}. Subsequent works, including BLIP-2, further demonstrated that lightweight trainable connectors can effectively align frozen visual encoders with large language models under a staged training pipeline \cite{li2023blip2}. Building on this line, visual instruction tuning methods such as LLaVA showed that instruction-following ability can be substantially improved by adapting models on diverse image--text instruction data \cite{liu2023visual}, while open-source systems such as Qwen-VL extended this paradigm toward more general multimodal understanding, grounding, and text reading \cite{bai2023qwenvl}. These studies provide the architectural and training foundation for our work. However, they mainly emphasize model design and overall data scaling, while paying less attention to how different categories of supervision should be temporally organized during instruction tuning.

\subsection{Specialized Supervision for Structured Reasoning and OCR-Centric Understanding}

A second line of work focuses on datasets and benchmarks that target specific multimodal capabilities beyond general-purpose visual instruction following. For structured reasoning, diagram and chart understanding benchmarks such as AI2D and ChartQA require models to reason over visual layouts, symbolic relations, and quantitative patterns, thereby testing abilities that are not fully captured by generic image description or question-answering data \cite{kembhavi2016diagram,masry2022chartqa}. For fine-grained perception, TextVQA and DocVQA highlight the importance of reading scene text, understanding document layouts, and integrating local textual evidence with global visual context \cite{singh2019textvqa,mathew2021docvqa}. Together, these benchmarks reveal that multimodal capability is inherently heterogeneous: a model that performs well on broad semantic understanding may still underperform on chart reasoning or OCR-intensive tasks. Our work is closely related to this line of research, but differs in emphasis. Instead of proposing a new benchmark or a new task-specific model, we investigate whether the organization of existing heterogeneous supervision sources affects the final balance among these capabilities.

\subsection{Data Organization, Curriculum Learning, and Continual Multimodal Adaptation}

Beyond model architecture and benchmark construction, an emerging line of research has begun to examine how data scheduling and sequential adaptation influence multimodal learning. Curriculum-oriented studies suggest that training order can affect optimization behavior and downstream capability acquisition, especially when supervision sources differ in complexity and modality composition \cite{bengio2009curriculum}. In parallel, recent work on continual or sequential multimodal instruction tuning has highlighted challenges such as task imbalance, catastrophic forgetting, and the tension between preserving general capability and acquiring specialized skills \cite{ge2025dmole,chen2025sefe}. These studies indicate that the temporal arrangement of multimodal supervision is not merely an implementation detail, but a factor that may shape both performance and stability. Nevertheless, prior work has typically focused on continual-learning algorithms, model-expansion strategies, or task-specific anti-forgetting mechanisms. Our study instead asks a more controlled question: under fixed trainable modules and approximately matched total exposure, how do curriculum training, direct mixture, balanced sampling, and reverse curriculum differ in balancing general understanding, structured reasoning, and OCR-centric perception?

\section{Methodology}

\subsection{Problem Formulation}

We study whether the \emph{organization of heterogeneous training data} affects the capability balance among three target dimensions in multimodal instruction tuning: general visual understanding, structured reasoning, and fine-grained OCR/document understanding. Let the training corpus be partitioned into four functional groups,
\begin{equation}
\mathcal{D} = \{D_0, D_1, D_2, D_3\},
\end{equation}
where $D_0$ denotes alignment data, $D_1$ denotes general visual instruction data, $D_2$ denotes structured reasoning data, and $D_3$ denotes OCR/document-centric data.

At training stage $s$, a dataset is first sampled from a stage-dependent categorical distribution $\pi_s(d)$, and then an instance $(x,y)$ is uniformly sampled from the selected dataset. The stage-wise training objective is
\begin{equation}
\mathcal{L}_s(\theta)
=
\mathbb{E}_{d \sim \pi_s}
\;
\mathbb{E}_{(x,y)\sim \mathcal{U}(d)}
\big[
\ell(f_{\theta}(x), y)
\big],
\end{equation}
where $\mathcal{U}(d)$ denotes uniform sampling within dataset $d$, $f_{\theta}$ is the multimodal model, and $\ell(\cdot)$ is the instruction-tuning loss.

Our goal is not to change the model architecture or optimization modules, but to examine whether different choices of $\pi_s$ across stages lead to different trade-offs among the target capabilities. This controlled formulation isolates data scheduling as the primary experimental variable.

\subsection{Capability-Oriented Data Partition}

We organize the training data according to their dominant functional roles in multimodal adaptation:
\begin{itemize}
    \item $D_0$: LLaVA-Pretrain 558K, used for initial visual--language alignment;
    \item $D_1$: ShareGPT4V, used for general semantic grounding and visual instruction following;
    \item $D_2$: AI2D and ChartQA, used for structured reasoning over diagrams, layouts, and charts;
    \item $D_3$: TextVQA and DocVQA, used for fine-grained text perception and document understanding.
\end{itemize}

This capability-oriented grouping serves two purposes. First, it makes explicit that different supervision sources contribute differently to downstream capability formation. Second, it enables us to define training schedules at the level of functional supervision rather than at the level of a single undifferentiated mixed corpus.

\subsection{Controlled Three-Stage Training}

All conditions share the same three-stage training protocol. We use a fixed set of trainable modules so that differences across runs can be attributed to data organization rather than model-side variation. Let $\Theta_P$ denote the projector parameters, $\Theta_L$ denote the LLM LoRA parameters, and $\Theta_V$ denote the vision tower parameters. The trainable parameter sets are
\begin{equation}
\begin{aligned}
\Theta_{\mathrm{train}}^{(1)} &= \{\Theta_P\}, \\
\Theta_{\mathrm{train}}^{(2,3)} &= \{\Theta_P, \Theta_L\}.
\end{aligned}
\end{equation}
Stage 1 performs initial visual--language alignment using $D_0$ only. Stages 2 and 3 perform post-alignment adaptation, where downstream task capabilities are injected under different data organization strategies. 
\

\subsection{Data Organization Strategies}

We compare four representative data organization strategies, denoted by $c \in \{A,B,C,D\}$. Each strategy is instantiated by a stage-specific sampling distribution $\pi_s^{(c)}(d)$.

\paragraph{Direct mixture (Condition A).}
This condition removes temporal curriculum and uses the same mixed sampling distribution across the post-alignment stages. It tests whether curriculum ordering provides benefits beyond exposure to the same overall supervision sources.

\paragraph{Curriculum training (Condition B).}
The main strategy adopts a coarse-to-specialized curriculum. Stage 2 emphasizes general visual instruction and structured reasoning, while Stage 3 expands the mixture to include OCR- and document-centric supervision. This design follows the intuition that broad semantic grounding and reasoning should be established before fine-grained textual perception is heavily emphasized.

\paragraph{Balanced sampling (Condition C).}
This condition assigns equal probability to all post-alignment task sources. It evaluates whether explicit task balancing leads to a better capability compromise than curriculum-style or naturally skewed mixtures.

\paragraph{Reverse curriculum (Condition D).}
This condition inverts the main curriculum by emphasizing OCR-heavy data earlier and shifting toward general instruction and reasoning later. It tests whether introducing fine-grained perceptual supervision too early is detrimental to the final capability balance.

To compare schedules more fairly, we approximately match the total exposure budget across conditions. Let $T_s$ denote the number of training steps in stage $s$. The expected exposure of dataset $d$ under condition $c$ is
\begin{equation}
E^{(c)}(d) = \sum_{s \in \{2,3\}} T_s \, \pi_s^{(c)}(d).
\end{equation}
Our comparison is designed such that the major difference across conditions lies in the \emph{temporal organization} of supervision rather than the total amount of supervision itself.

\subsection{Training Dynamics Analysis}

Beyond final validation scores, we analyze training dynamics from two complementary perspectives: capability evolution under periodic evaluation and optimization stability reflected by the training loss trajectory. This design allows us to examine not only where a model converges, but also how its different capabilities emerge and how stable the optimization process remains throughout training.

\paragraph{Stage-wise Capability Evaluation.}
At fixed training-step intervals, we evaluate the model on four capability categories: general capability, reasoning capability, fine-grained detail capability, and overall capability. Let the evaluation steps be denoted by $\mathcal{T}_{\mathrm{eval}}=\{t_1,t_2,\dots\}$. 
 This stage-wise protocol enables us to characterize how different data organization strategies shape the trajectory of capability acquisition during training, rather than only comparing their final converged checkpoints.

\paragraph{Loss Curve and Local Fluctuation.}
In parallel, we record the training loss at every optimization step. Let $\ell_t$ denote the training loss at step $t$. To characterize short-term optimization stability, we define a local fluctuation measure over a window of size $w$ as
\begin{equation}
\sigma_t^{(\ell)}
=
\mathrm{Std}\big(\ell_{t-w+1}, \ldots, \ell_t\big).
\end{equation}

\paragraph{Spike Frequency}
To quantify abrupt instability during training, we compute the local mean and local standard deviation of the loss within a sliding window of size $u$. Specifically, at training step $t$, the local mean is defined as
\begin{equation}
\mu_t^{(\ell)}
=
\frac{1}{u}
\sum_{i=t-u+1}^{t} \ell_i,
\end{equation}
and the local standard deviation is defined as
\begin{equation}
\sigma_t^{(\ell)}
=
\mathrm{Std}\!\left(
\ell_{t-u+1}, \ldots, \ell_t
\right).
\end{equation}

where $\mathbb{I}[\cdot]$ denotes the indicator function. A training step is regarded as a spike if its loss is significantly higher or lower than the local mean by more than two local standard deviations.

Let the total number of training steps be $T$, and let the sliding-window size be $u$. Then the total number of valid windows is
\begin{equation}
N_{\mathrm{window}} = T - u + 1.
\end{equation}
For each window ending at step $t$ (where $t=u,u+1,\ldots,T$), we regard the window as containing a spike if the most recent loss value $\ell_t$ satisfies
\begin{equation}
\ell_t > \mu_t^{(\ell)} + 2\sigma_t^{(\ell)}
\quad \text{or} \quad
\ell_t < \mu_t^{(\ell)} - 2\sigma_t^{(\ell)}.
\end{equation}
Accordingly, the spike indicator is defined as
\begin{equation}
\mathbb{I}_t^{\mathrm{spike}}
=
\mathbb{I}\!\left[
\ell_t > \mu_t^{(\ell)} + 2\sigma_t^{(\ell)}
\;\;\text{or}\;\;
\ell_t < \mu_t^{(\ell)} - 2\sigma_t^{(\ell)}
\right],
\end{equation}
where $\mathbb{I}[\cdot]$ denotes the indicator function.

The spike frequency is then defined as
\begin{equation}
f_{\mathrm{spike}}
=
\frac{1}{N_{\mathrm{window}}}
\sum_{t=u}^{T} \mathbb{I}_t^{\mathrm{spike}}
=
\frac{1}{T-u+1}
\sum_{t=u}^{T} \mathbb{I}_t^{\mathrm{spike}}.
\end{equation}
This metric represents the proportion of valid sliding windows in which the most recent loss exceeds the threshold.

\paragraph{Stage Transition Stability.}

Since our training protocol is stage-based, we further examine whether switching from an old stage to a new stage introduces significant optimization disturbance. Let the stage transition occur at step $t_s$, where $t_s^{-}$ and $t_s^{+}$ denote the nearest logged training steps immediately before and after the transition, respectively. To measure the relative change in loss across the stage boundary, we define the stage-transition ratio as
\begin{equation}
r_{\mathrm{stage}}
=
\frac{\ell_{t_s^{+}} - \ell_{t_s^{-}}}{\ell_{t_s^{-}}}.
\end{equation}
Here, $\ell_{t_s^{-}}$ denotes the loss at the end of the old stage, and $\ell_{t_s^{+}}$ denotes the loss after entering the new stage.

\subsection{Evaluation Metrics}

We evaluate model performance on five validation tasks spanning the target capability spectrum: \textsc{General-Val}, \textsc{AI2D}, \textsc{ChartQA}, \textsc{TextVQA}, and \textsc{DocVQA}. In addition to reporting each task score individually, we define aggregated metrics to characterize capability balance across different task families. Specifically, for a given evaluation checkpoint, we define
\begin{equation}
\mathrm{Reasoning} = \frac{\mathrm{AI2D} + \mathrm{ChartQA}}{2},
\end{equation}
\begin{equation}
\mathrm{Detail} = \frac{\mathrm{TextVQA} + \mathrm{DocVQA}}{2},
\end{equation}
\begin{equation}
\mathrm{Overall} = \frac{\mathrm{General\text{-}Val} + \mathrm{AI2D} + \mathrm{ChartQA} + \mathrm{TextVQA} + \mathrm{DocVQA}}{5}.
\end{equation}
Here, $\mathrm{Reasoning}$ summarizes structured reasoning performance, $\mathrm{Detail}$ captures fine-grained OCR- and document-oriented capability, and $\mathrm{Overall}$ measures the average validation performance across the full capability spectrum. These metrics allow us to assess not only final average performance, but also how different training schedules allocate modeling capacity across general, reasoning-oriented, and detail-oriented tasks.

\section{Experimental}
\subsection{Experimental Setup}

We follow the controlled three-stage training protocol described in Section 3. Across all conditions, the trainable modules, optimization pipeline, and model backbone remain fixed; the only varying factor is the organization of post-alignment training data. Stage~1 uses $D_0$ for initial visual--language alignment with projector-only training. Stages~2 and~3 jointly optimize the projector and LLM LoRA while keeping the vision tower frozen throughout.

For training, we use all available samples from each dataset. For evaluation, we use a sampled subset of the evaluation set. 

We compare four data organization strategies: direct mixture (A), curriculum training (B), balanced sampling (C), and reverse curriculum (D). Table~\ref{tab:setup} summarizes the stage-wise data composition of all conditions. The main comparison focuses on whether curriculum order improves the balance among general understanding, structured reasoning, and OCR-centric perception.

\begin{table*}[t]
\centering
\small
\caption{Summary of stage-wise data organization strategies. S4V denotes ShareGPT4V, TXT denotes TextVQA, and DOC denotes DocVQA.}
\label{tab:setup}
\resizebox{\textwidth}{!}{
\begin{tabular}{l l l l l}
\hline
Condition & Strategy & Stage 1 & Stage 2 & Stage 3 \\
\hline
A & Direct mixture &
$D_0$ only &
S4V: 0.50, AI2D: 0.13, ChartQA: 0.13, TXT: 0.12, DOC: 0.12 &
S4V: 0.50, AI2D: 0.13, ChartQA: 0.13, TXT: 0.12, DOC: 0.12 \\
B & Curriculum &
$D_0$ only &
S4V: 0.70, AI2D: 0.15, ChartQA: 0.15 &
S4V: 0.20, AI2D: 0.10, ChartQA: 0.10, TXT: 0.30, DOC: 0.30 \\
C & Balanced sampling &
$D_0$ only &
S4V: 0.20, AI2D: 0.20, ChartQA: 0.20, TXT: 0.20, DOC: 0.20 &
S4V: 0.20, AI2D: 0.20, ChartQA: 0.20, TXT: 0.20, DOC: 0.20 \\
D & Reverse curriculum &
$D_0$ only &
S4V: 0.20, AI2D: 0.10, ChartQA: 0.10, TXT: 0.30, DOC: 0.30 &
S4V: 0.70, AI2D: 0.15, ChartQA: 0.15 \\
\hline
\end{tabular}
}
\end{table*}

\subsection{Main Results}
Table~\ref{tab:main_results_grouped} summarizes the main validation results under different data-organization strategies. Overall, the curriculum design (Condition~B) achieves the best \textit{Overall} score (73.7), outperforming direct mixture (Condition~A, 72.1), balanced sampling (Condition~C, 72.8), and reverse curriculum (Condition~D, 71.2). This suggests that a staged progression that prioritizes general understanding and structured reasoning before introducing OCR-intensive supervision yields the most favorable overall capability trade-off. More specifically, Condition~B attains the best performance on \textit{General-Val} (73.4), \textit{AI2D} (76.0), and \textit{ChartQA} (74.6), and correspondingly achieves the highest \textit{Reasoning} average (75.3), indicating that this training order is particularly beneficial for developing general multimodal understanding and structured reasoning.

In contrast, balanced sampling (Condition~C) performs best on the OCR-oriented tasks, achieving the highest scores on \textit{TextVQA} (72.6) and \textit{DocVQA} (73.5), which leads to the strongest \textit{OCR} average (73.1). This pattern suggests that more uniform task exposure is advantageous for fine-grained text perception. However, Condition~C consistently trails Condition~B on \textit{General-Val}, \textit{AI2D}, and \textit{ChartQA}, and therefore does not surpass the curriculum strategy in overall performance. Direct mixture (Condition~A) serves as a reasonable baseline, but it shows no clear advantage on any metric, indicating that exposure to the same supervision sources alone is insufficient; the organization of that supervision also matters. Reverse curriculum (Condition~D) yields the weakest \textit{Overall} performance and underperforms substantially on \textit{General-Val}, \textit{AI2D}, and \textit{ChartQA}. Although its OCR-related scores are slightly higher than those of Condition~A, these gains are insufficient to compensate for the loss in broader understanding and reasoning ability. Taken together, these results indicate that emphasizing OCR-heavy supervision too early may interfere with the acquisition of more general multimodal capabilities.

\begin{table*}[t]
\centering
\small
\setlength{\tabcolsep}{5pt}
\renewcommand{\arraystretch}{1.15}
\caption{ Main results on the validation benchmarks under the proposed data-organization scheme. Higher is better. The best score in each column is \textbf{bolded}.}
\label{tab:main_results_grouped}
\resizebox{\textwidth}{!}{
\begin{tabular}{l|c|ccc|ccc|c}
\toprule
\multirow{2}{*}{Condition}
& \multicolumn{1}{c|}{General}
& \multicolumn{3}{c|}{Reasoning}
& \multicolumn{3}{c|}{OCR}
& \multirow{2}{*}{Overall} \\
\cmidrule(lr){2-2} \cmidrule(lr){3-5} \cmidrule(lr){6-8}
& General-Val
& AI2D & ChartQA & Reasoning
& TextVQA & DocVQA & OCR
& \\
\midrule
A
& 72.1 
& 74.0  & 72.0  & 73.0 
& 70.8  & 71.4  & 71.1 
& 72.1  \\
B
& \textbf{73.4 }
& \textbf{76.0 } & \textbf{74.6 } & \textbf{75.3 }
& 71.8  & 72.9  & 72.4 
& \textbf{73.7 } \\
C
& 71.7 
& 73.3  & 72.9  & 73.1 
& \textbf{72.6 } & \textbf{73.5 } & \textbf{73.1 }
& 72.8  \\
D
& 69.8 
& 72.1  & 70.9  & 71.5 
& 71.4  & 71.8  & 71.6 
& 71.2  \\
\bottomrule
\end{tabular}
}
\end{table*}
\subsection{Capability Trade-off Analysis}

\begin{figure}[t]
    \centering
    \includegraphics[width=0.8\linewidth]{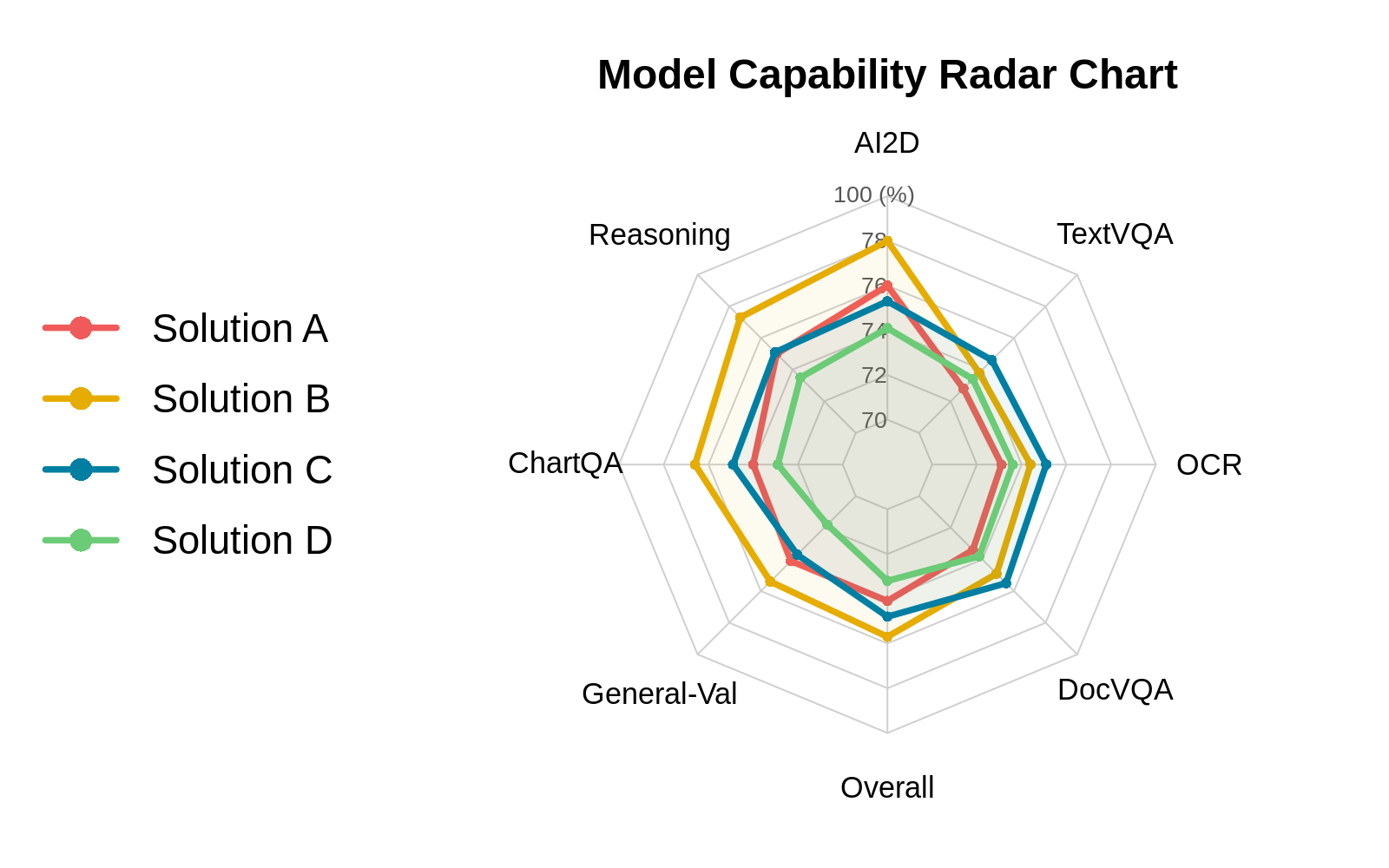}
    \caption{Comparison of all the scores across different training conditions.}
    \label{fig:tradeoff}
\end{figure}

To provide a more intuitive view of the trade-offs across capability dimensions, Figure~\ref{fig:tradeoff} visualizes the performance profile of each training condition as a radar chart over multiple validation axes. A clear pattern emerges: the curriculum design (Condition~B) exhibits the most balanced and consistently expanded profile. It dominates on \textit{AI2D}, \textit{ChartQA}, \textit{Reasoning}, \textit{General-Val}, and \textit{Overall}, indicating that this strategy not only yields the strongest aggregate performance but also better supports the joint development of general understanding and structured reasoning. In contrast, balanced sampling (Condition~C) shows a more pronounced advantage along the \textit{TextVQA}, \textit{DocVQA}, and aggregated \textit{OCR} axes, suggesting that it shifts the model toward stronger text-centric perception, albeit at the cost of slightly weaker reasoning and overall generalization than Condition~B.

Direct mixture (Condition~A) presents a relatively stable profile across dimensions, but its contour remains consistently inside that of Condition~B, further suggesting that merely combining the same supervision sources is insufficient for optimal performance; the organization of supervision also matters. Reverse curriculum (Condition~D) forms the smallest overall contour on most axes, with particularly noticeable degradation on \textit{General-Val}, \textit{ChartQA}, and \textit{Reasoning}. This suggests that introducing OCR-heavy supervision too early may hinder the acquisition of broader multimodal understanding. Overall, Figure~\ref{fig:tradeoff} corroborates the findings in Table~\ref{tab:main_results_grouped}: an appropriate curriculum-based data organization leads to a better global balance across capability dimensions, rather than producing gains that are confined to a single sub-skill.

\subsection{Training Dynamics and Efficiency}
\begin{figure}[t]
    \centering
    \includegraphics[width=0.8\linewidth]{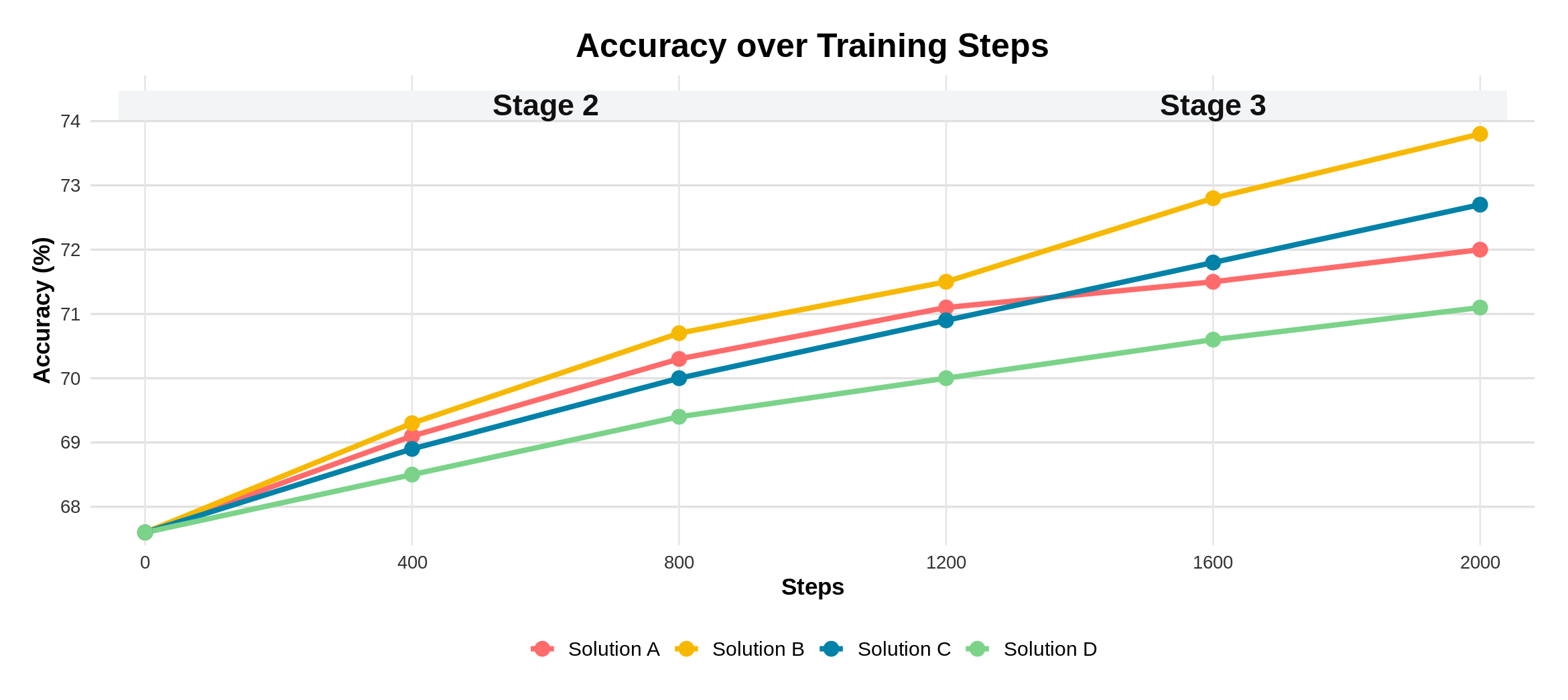}
    \caption{Comparison of \textit{Overall} across different training conditions.}
    \label{fig:convergence_curve}
\end{figure}

Beyond final validation performance, we further examine whether the advantage of curriculum-style data organization is also reflected in optimization dynamics and computational efficiency. Following the metrics defined in Section~3.5, we analyze convergence behavior, short-term loss fluctuation, abrupt instability. The results are summarized in Table~\ref{tab:dynamics_efficiency}, while Figure~\ref{fig:convergence_curve} visualizes the evolution of the \textit{Overall} score over training steps.

A clear difference emerges in convergence behavior. Condition~B reaches a high-performing regime earlier than the other strategies and maintains the strongest \textit{Overall} trajectory throughout most of post-alignment training, indicating that curriculum ordering improves not only the final capability trade-off but also the efficiency with which useful capability is acquired. By contrast, Condition~D converges the slowest and remains consistently below the other conditions, suggesting that introducing OCR-heavy supervision too early leads to a less effective optimization path. Conditions~A and~C exhibit intermediate behavior: both improve steadily, but neither matches the convergence speed or final plateau achieved by Condition~B.

The stability statistics further support this pattern. Condition~B yields the smallest loss fluctuation and the fewest instability spikes. This result is consistent with the intuition behind the proposed curriculum: establishing broad semantic grounding and structured reasoning before emphasizing OCR-intensive supervision can reduce optimization conflict across heterogeneous tasks. In contrast, Condition~D exhibits the largest fluctuation statistics and the highest spike count, suggesting that reverse curriculum introduces a more unstable learning process. Notably, the stage-transition analysis provides an additional nuance: because Conditions~A and~C keep their mixtures unchanged across post-alignment stages, they are expected to show relatively smaller transition disturbance, whereas Condition~B incurs a moderate transition cost when OCR/document-centric supervision is deliberately introduced in the later stage. Nevertheless, this temporary disturbance is compensated by faster convergence and better final performance.

\begin{table}[t]
\centering
\small
\caption{Training dynamics and efficiency statistics. Lower is better for loss standard deviation, spike frequency, stage-transition loss gap. }
\label{tab:dynamics_efficiency}
\begin{tabular}{lccc}
\toprule
Condition & Loss Std$\downarrow$ & Spike Freq.$\downarrow$ & Stage Tran Stab   \\
\midrule
A & 0.117 & 2.12\% & \textbf{0.29} \\
B & \textbf{0.096} & \textbf{1.48\%} & 0.37 \\
C & 0.109 & 1.84\% & 0.27 \\
D & 0.136 & 2.73\% & 0.68 \\
\bottomrule
\end{tabular}
\end{table}

\subsection{Discussion}
Taken together, the results indicate that data organization is a first-order design variable in multimodal instruction tuning rather than a secondary implementation choice. Under the controlled setting of this work, where the trainable modules and approximate total exposure are matched across conditions, changing only the temporal arrangement of supervision is already sufficient to produce markedly different capability profiles. In particular, curriculum training yields the most favorable global trade-off, balanced sampling remains attractive when OCR-centric performance is prioritized, and reverse curriculum is consistently less effective in balancing broad understanding, reasoning, and fine-grained perception.

A plausible interpretation is that the acquisition order of heterogeneous supervision affects how limited adaptation capacity is allocated during post-alignment tuning. Emphasizing general instruction following and structured reasoning earlier appears to establish a stronger semantic and relational foundation, upon which OCR- and document-centric capabilities can later be incorporated without severely disrupting previously acquired competencies. In contrast, placing OCR-heavy supervision too early may bias optimization toward local textual and perceptual patterns before broader multimodal abstractions are sufficiently consolidated, which in turn weakens subsequent generalization and reasoning. This interpretation is also consistent with the training-dynamics observations, where the curriculum schedule exhibits smoother optimization and faster convergence than the alternative schedules.

At the same time, the present findings should be interpreted within the scope of the current experimental design. Our conclusions are drawn under a fixed three-stage training pipeline with a frozen vision backbone and projector+LoRA adaptation, and are evaluated on a validation suite centered on general understanding, reasoning, and OCR/document tasks. Accordingly, the results should not be overgeneralized as evidence that a single curriculum is universally optimal for all model scales, backbones, or downstream objectives. For example, when the deployment goal places disproportionate weight on OCR-intensive capability, a more balanced sampling strategy may still be preferable in practice. More broadly, our study suggests that data scheduling should be treated as an explicit optimization dimension in multimodal adaptation, and it motivates future work on adaptive curricula, finer-grained task grouping, and schedule design under broader model and benchmark settings.

\section{Conclusion}

In this work, we investigated whether the organization of heterogeneous training data affects the capability balance of multimodal instruction tuning. Under a controlled setting with fixed trainable modules and approximately matched data exposure, we compared four representative strategies: direct mixture, curriculum training, balanced sampling, and reverse curriculum. Our results suggest that data scheduling is not merely an implementation detail, but a key factor that shapes the trade-off among general visual understanding, structured reasoning, and fine-grained OCR/document understanding.

Among the compared strategies, curriculum-style training yields the most favorable overall balance. By first consolidating broad semantic grounding and structured reasoning, and then introducing OCR- and document-intensive supervision, this strategy achieves stronger overall performance while maintaining competitive OCR capability. In contrast, direct mixture provides a reasonable but weaker baseline, balanced sampling tends to favor OCR-oriented tasks at the expense of broader capability balance, and reverse curriculum is generally less effective in both performance and optimization stability. These findings indicate that the temporal organization of supervision can substantially influence both final capability allocation and training dynamics.

More broadly, our study highlights data organization as an explicit design dimension in multimodal model adaptation. Future work may extend this direction by exploring adaptive or performance-aware scheduling strategies, more fine-grained task grouping, and broader model scales and benchmark settings. We hope these findings can provide practical guidance for building multimodal systems that require a careful balance among general understanding, reasoning, and OCR-intensive perception.


{\small
\begin{thebibliography}{99}

\bibitem[Liu et~al.(2023)]{liu2023visual}
Liu, H., Li, C., Wu, Q., and Lee, Y.~J. (2023).
Visual Instruction Tuning.
\textit{arXiv preprint arXiv:2304.08485}.

\bibitem[Dai et~al.(2023)]{dai2023instructblip}
Dai, W., Li, J., Li, D., Tiong, A.~M.~H., Zhao, J., Wang, W., Li, B., Fung, P., and Hoi, S. (2023).
InstructBLIP: Towards General-purpose Vision-Language Models with Instruction Tuning.
\textit{arXiv preprint arXiv:2305.06500}.

\bibitem[Hu et~al.(2023a)]{hu2023paperowl}
Hu, A., Shi, Y., Xu, H., Ye, J., Ye, Q., Yan, M., Li, C., Qian, Q., Zhang, J., and Huang, F. (2023).
mPLUG-PaperOwl: Scientific Diagram Analysis with the Multimodal Large Language Model.
\textit{arXiv preprint arXiv:2311.18248}.

\bibitem[Xia et~al.(2024)]{xia2024chartx}
Xia, R., Zhang, B., Ye, H., Yan, X., Liu, Q., Zhou, H., Chen, Z., Ye, P., Dou, M., Shi, B., Yan, J., and Qiao, Y. (2024).
ChartX \& ChartVLM: A Versatile Benchmark and Foundation Model for Complicated Chart Reasoning.
\textit{arXiv preprint arXiv:2402.12185}.

\bibitem[Hu et~al.(2024)]{hu2024mplug}
Hu, A., Xu, H., Ye, J., Yan, M., Zhang, L., Zhang, B., Zhang, J., Jin, Q., Huang, F., and Zhou, J. (2024).
mPLUG-DocOwl 1.5: Unified Structure Learning for OCR-free Document Understanding.
In \textit{Findings of the Association for Computational Linguistics: EMNLP 2024}, pages 3096--3120.

\bibitem[Li et~al.(2023)]{li2023blip2}
Li, J., Li, D., Savarese, S., and Hoi, S. (2023).
BLIP-2: Bootstrapping Language-Image Pre-training with Frozen Image Encoders and Large Language Models.
In \textit{Proceedings of the 40th International Conference on Machine Learning}, pages 19730--19742.

\bibitem[Kembhavi et~al.(2016)]{kembhavi2016diagram}
Kembhavi, A., Salvato, M., Kolve, E., Seo, M., Hajishirzi, H., and Farhadi, A. (2016).
A Diagram Is Worth a Dozen Images.
In \textit{Computer Vision -- ECCV 2016}, pages 235--251. Cham: Springer.

\bibitem[Masry et~al.(2022)]{masry2022chartqa}
Masry, A., Long, D.~X., Tan, J.~Q., Joty, S., and Hoque, E. (2022).
ChartQA: A Benchmark for Question Answering about Charts with Visual and Logical Reasoning.
In \textit{Findings of the Association for Computational Linguistics: ACL 2022}, pages 2263--2279.

\bibitem[Singh et~al.(2019)]{singh2019textvqa}
Singh, A., Natarajan, V., Shah, M., Jiang, Y., Chen, X., Batra, D., Parikh, D., and Rohrbach, M. (2019).
Towards VQA Models That Can Read.
In \textit{Proceedings of the IEEE/CVF Conference on Computer Vision and Pattern Recognition (CVPR)}, pages 8317--8326.

\bibitem[Mathew et~al.(2021)]{mathew2021docvqa}
Mathew, M., Karatzas, D., and Jawahar, C.~V. (2021).
DocVQA: A Dataset for VQA on Document Images.
In \textit{Proceedings of the IEEE/CVF Winter Conference on Applications of Computer Vision (WACV)}, pages 2200--2209.

\bibitem[Hu et~al.(2022)]{hu2022lora}
Hu, E.~J., Shen, Y., Wallis, P., Allen-Zhu, Z., Li, Y., Wang, S., Wang, L., and Chen, W. (2022).
LoRA: Low-Rank Adaptation of Large Language Models.
In \textit{International Conference on Learning Representations}.

\bibitem[Ge et~al.(2025)]{ge2025dmole}
Ge, C., Wang, X., Zhang, Z., Chen, H., Fan, J., Huang, L., Xue, H., and Zhu, W. (2025).
Dynamic Mixture of Curriculum LoRA Experts for Continual Multimodal Instruction Tuning.
In \textit{Proceedings of the 42nd International Conference on Machine Learning}, \textit{Proceedings of Machine Learning Research} \textbf{267}, pages 19011--19033. PMLR.

\bibitem[Chen et~al.(2025)]{chen2025sefe}
Chen, J., Cong, R., Zhao, Y., Yang, H., Hu, G., Ip, H., and Kwong, S. (2025).
SEFE: Superficial and Essential Forgetting Eliminator for Multimodal Continual Instruction Tuning.
In \textit{Proceedings of the 42nd International Conference on Machine Learning}, \textit{Proceedings of Machine Learning Research} \textbf{267}, pages 7982--8001. PMLR.

\bibitem[Bengio et~al.(2009)]{bengio2009curriculum}
Bengio, Y., Louradour, J., Collobert, R., and Weston, J. (2009).
Curriculum Learning.
In \textit{Proceedings of the 26th International Conference on Machine Learning}, pages 41--48. ACM.

\bibitem[Alayrac et~al.(2022)]{alayrac2022flamingo}
Alayrac, J.-B., Donahue, J., Luc, P., Miech, A., Barr, I., Hasson, Y., Lenc, K., Mensch, A., Millican, K., Reynolds, M., Ring, R., Rutherford, E., Cabi, S., Han, T., Gong, Z., Samangooei, S., Monteiro, M., Menick, J., Borgeaud, S., Brock, A., Nematzadeh, A., Sharifzadeh, S., Bi\'nkowski, M., Barreira, R., Vinyals, O., Zisserman, A., and Simonyan, K. (2022).
Flamingo: A Visual Language Model for Few-Shot Learning.
In \textit{Advances in Neural Information Processing Systems 35}, pages 23716--23736.

\bibitem[Bai et~al.(2023)]{bai2023qwenvl}
Bai, J., Bai, S., Yang, S., Wang, S., Tan, S., Wang, P., Lin, J., Zhou, C., and Zhou, J. (2023).
Qwen-VL: A Versatile Vision-Language Model for Understanding, Localization, Text Reading, and Beyond.
\textit{arXiv preprint arXiv:2308.12966}.

\end{thebibliography}

}
\end{document}